\documentclass{article}
\usepackage{ijcai23}
\usepackage{times}
\usepackage{soul}
\usepackage{url}
\usepackage[hidelinks]{hyperref}
\usepackage[utf8]{inputenc}
\usepackage[small]{caption}
\usepackage{graphicx}
\usepackage{amsmath}
\usepackage{amsthm}
\usepackage{booktabs}
\usepackage{algorithm}
\usepackage{algorithmic}
\usepackage{amsfonts,amssymb}
\usepackage{dsserif}
\usepackage{multirow}
\usepackage{bbding}
\usepackage[switch]{lineno}

\title{Linguistic More: Taking a Further Step toward Efficient and Accurate Scene Text Recognition}

\author{
Boqiang Zhang
\and
Hongtao Xie\footnote{Corresponding Author}\and
Yuxin Wang\and
Jianjun Xu\and
Yongdong Zhang
\affiliations
University of Science and Technology of China, Hefei, China\\
\emails
\{cyril, wangyx58, xujj1998\}@mail.ustc.edu.cn,
\{htxie, zhyd73\}@ustc.edu.cn
}

\begin{document}
\maketitle
\begin{abstract}
    Vision model have gained increasing attention due to their simplicity and efficiency in Scene Text Recognition (STR) task. However, due to lacking the perception of linguistic knowledge and information, recent vision models suffer from two problems: (1) the pure vision-based query results in attention drift, which usually causes poor recognition and is summarized as linguistic insensitive drift (LID) problem in this paper. (2) the visual feature is suboptimal for the recognition in some vision-missing cases (e.g. occlusion, etc.). To address these issues, we propose a \textbf{L}inguistic \textbf{P}erception \textbf{V}ision model (LPV), which explores the linguistic capability of vision model for accurate text recognition. To alleviate the LID problem, we introduce a Cascade Position Attention (CPA) mechanism that obtains high-quality and accurate attention maps through step-wise optimization and linguistic information mining. Furthermore, a Global Linguistic Reconstruction Module (GLRM) is proposed to improve the representation of visual features by perceiving the linguistic information in the visual space, which gradually converts visual features into semantically rich ones during the cascade process. Different from previous methods, our method obtains SOTA results while keeping low complexity (92.4\% accuracy with only 8.11M parameters). Code is available at \href{https://github.com/CyrilSterling/LPV}{https://github.com/CyrilSterling/LPV}.
\end{abstract}

\begin{figure}[!t]
     \centering
     \includegraphics[scale=0.5]{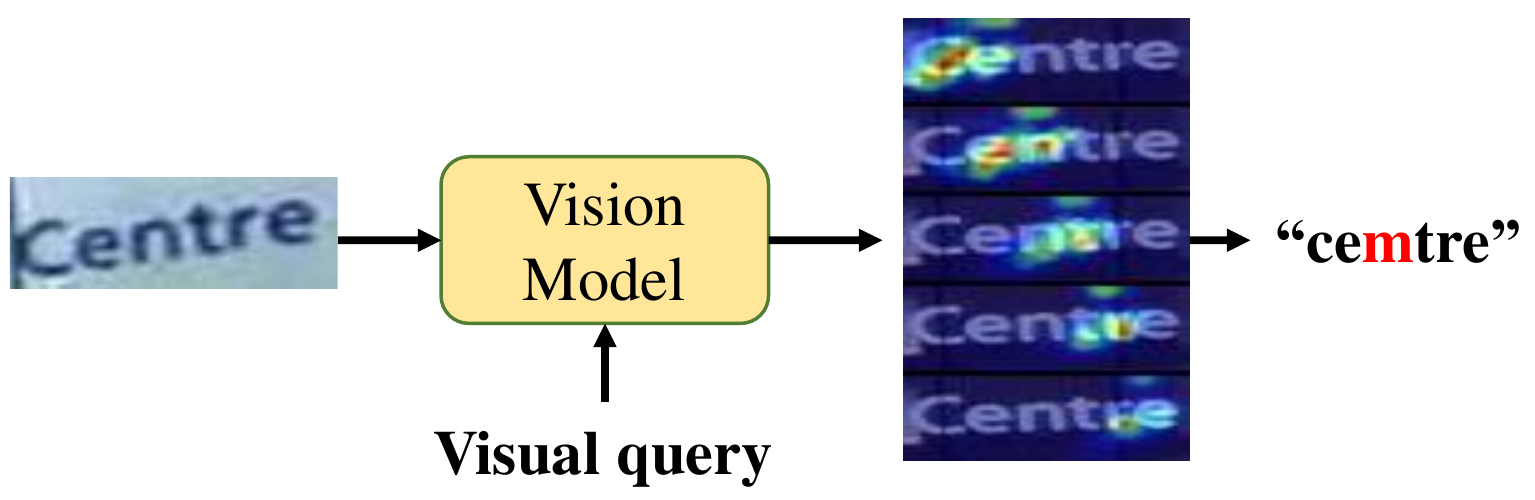}
     
     (a)
     
     \includegraphics[scale=0.35]{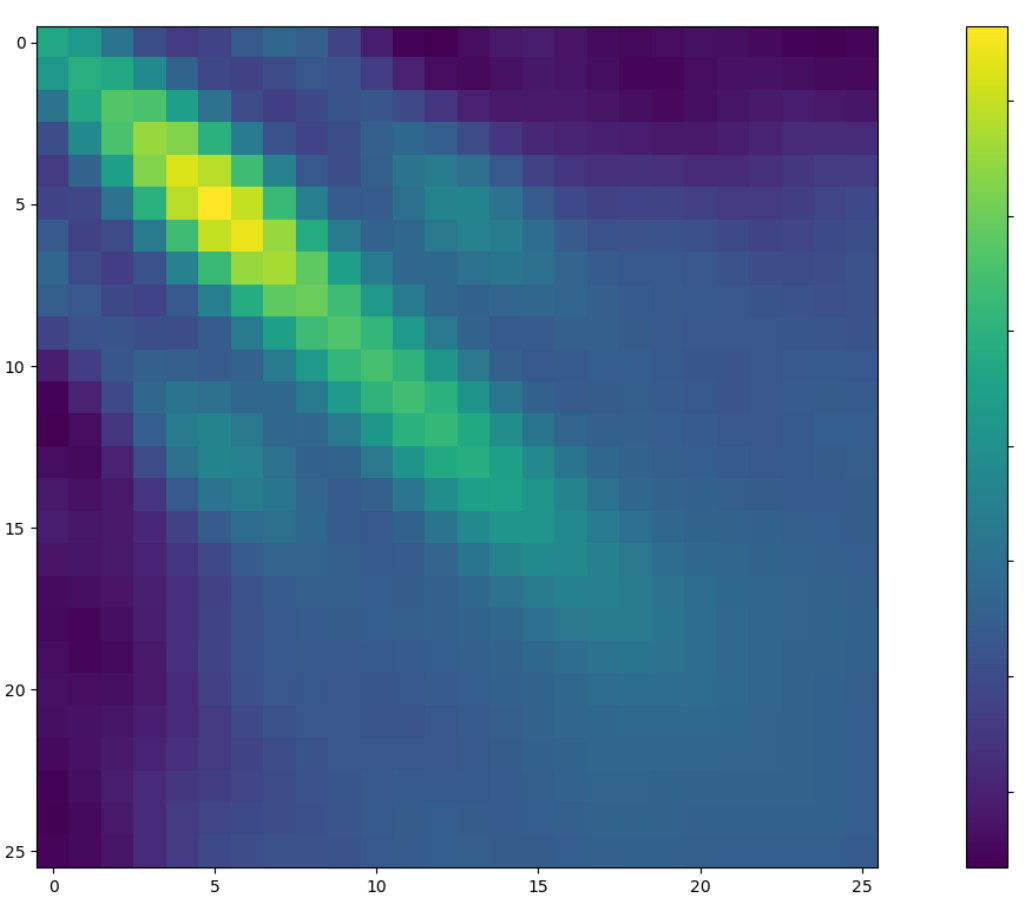}

     (b)
     \caption{Our motivation. (a) The basic structure of previous vision-only models with an attention-based decoder and a visual query to decode the characters, which has the problem of attention drift. (b) The visualization of the dot similarity between the query vectors at each position of ABINet~\protect\cite{abinet}.}
     \label{fig:motivation}
\end{figure}

\section{Introduction}\label{sec:introduction}
Scene Text Recognition (STR) is a meaningful task in computer vision that aims to understand the textual information from the cropped image of natural scenes ~\cite{long2021scene,shi2016end,abinet,jianjun,chen1}. Due to the lack of language modal information of other perception tasks, STR is widely used in Visual Questions and Answers (VQA), automatic pilots, {\em etc}.

Early work has generally treated STR as a visual task, using an encoder to get visual features and, after sequence modeling, a CTC-based~\cite{CTC} or attention-based decoder to obtain the predicted characters. The attention-based decoder uses a visual query to decode the position of each character, so it is accurate for arbitrary shape text recognition. In addition, due to its simplicity and effectiveness, the attention-based decoder is currently the mainstream solution for vision model~\cite{visionlan,abinet,MGP,chen2,wang2022petr}. Such methods have a simple structure that can be efficient in most application scenarios. Though these vision-only methods have achieved promising results, there are still two problems.

The first problem is the attention drift in the attention-based decoders, which is not received enough attention in recent researches. Attention drift is when the area of attention region is not aligned with the target character (Figure. \ref{fig:motivation} (a)). RobustScanner~\cite{robustscanner} deeply analyzed attention drift and proved that the query vectors in the decoder encode not only context but also positional information. However, this positional information is easily drowned out by the introduction of other information. Note that recent methods use a pure vision-based query, which is fixed when inputting different images. We further visualize the dot similarity between the query vectors at each position in ABINet~\cite{abinet}. As shown in Figure. \ref{fig:motivation} (b), it is observed that the query at each position is similar to that at neighboring positions. This will lead to similar features when decoding the attention map of neighboring characters, thereby causing attention drift. Therefore, we indicate that the attention drift comes from decoding different images with the fixed vision-based query, which is linguistic insensitive. We summarize this issue as the Linguistic Insensitive Drift (LID) problem. Thus, how to eliminate the LID issue and obtain an accurate attention map is the key for robust text recognition.

Another problem is that visual feature is suboptimal for recognition in some vision-missing cases. To solve this problem, recent methods introduce the linguistic knowledge to assist the vision model. However, it is hard for vision model to obtain linguistic information efficiently and accurately. VisionLAN~\cite{visionlan} designed a masked language-aware module to randomly occlude a character in the training stage which guides the vision model to utilize the linguistic information in text images. But the model introduces additional modules and requires separate pre-training. MGP-STR~\cite{MGP} proposed a multi-granularity prediction strategy to inject information from the language modality into the model in an implicit way. However, this network requires a huge number of parameters. Thus, how to perceive linguistic information with an efficient structure and a simple training strategy is a great challenge for text recognition.

\begin{figure*}[!t]
\centering
\includegraphics[scale=0.48]{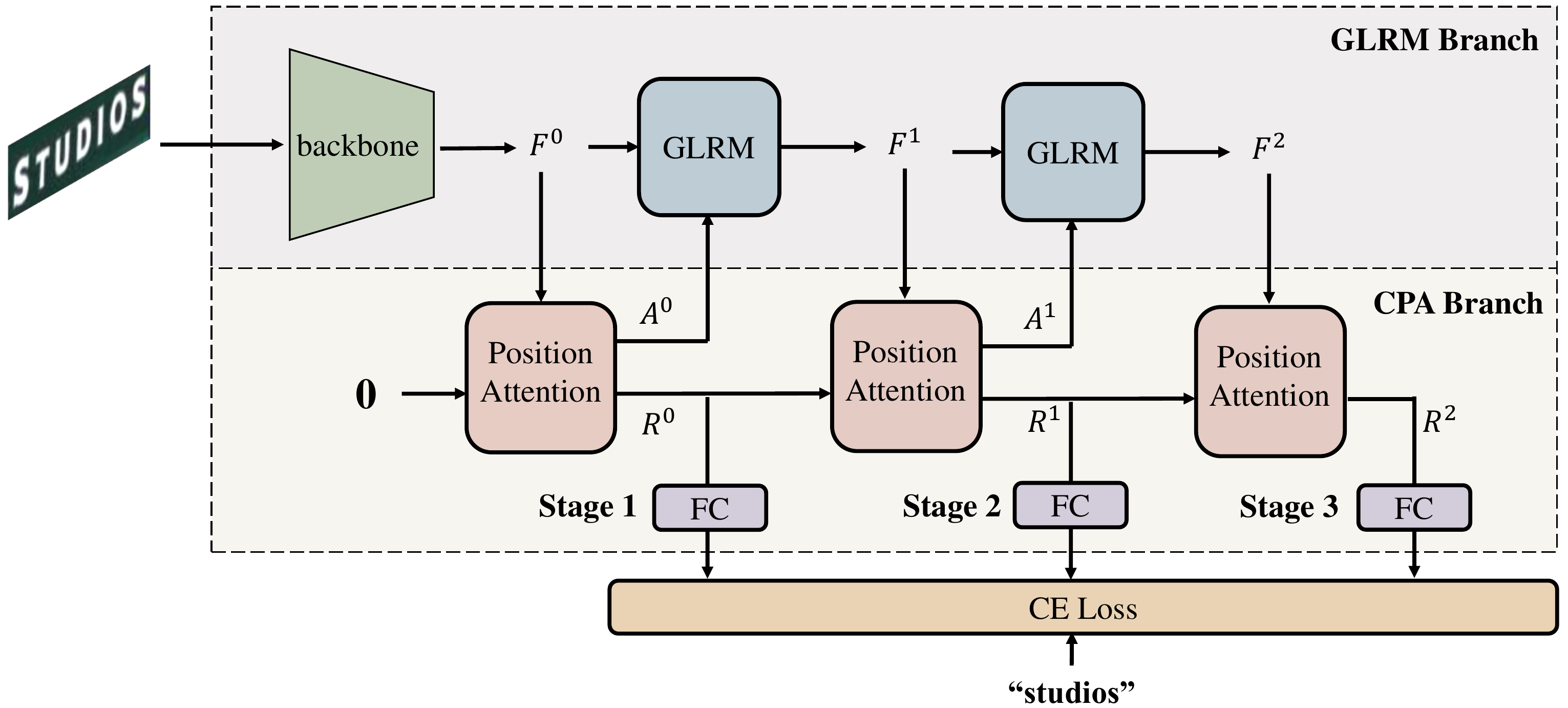}
\caption{The pipeline of our LPV. The pipeline mainly contains two branches: GLRM branch and CPA branch. GLRM branch continuously enhances the feature using proposed GLRM. CPA branch takes the feature as input and hierarchically decodes the attention map and feature of each character. CE Loss means the cross-entropy loss.}
\label{fig:pipeline}
\end{figure*}

To enhance the linguistic perception of both query and feature in a simple way, we propose a concise Linguistic Perception Vision model (LPV). The pipeline of our LPV is shown in Figure.~\ref{fig:pipeline}. The pipeline mainly consists of two parts: the GLRM branch and the CPA branch. The GLRM branch continuously enhances the features of the input image. In this branch, the visual features $\mathbf{F}^0$ are firstly extracted from the backbone. Then, Global Linguistic Reconstruction Module (GLRM) enhances the features of the previous stage $\mathbf{F}^{i-1}$ into the features of the current stage $\mathbf{F}^i$ using the mask generated by the attention map $\mathbf{A}^{i-1}$. In this way, linguistic information is aggregated in GLRM and the visual features can be gradually transformed into semantic-rich features. Meanwhile, the GLRM ensures the simplification of the pipeline and there are no redundant modules. The CPA branch hierarchically optimizes the attention map and the query using the cascade position attention mechanism. Each Position Attention Module (PAM) takes the visual features $\mathbf{F}^i$ as input, and obtains the attention map $\mathbf{A}^i$ and the features $\mathbf{R}^i$ of each character. Note that the prior query of the first PAM is initialized to $\boldsymbol{0}$, which means we have no prior to each character. PAM at $i^{th}$ stage uses $\mathbf{R}^{i-1}$ as the prior query. Such an operation can take the recognition result of the previous stage as a priori and re-perform the similarity calculation for the enhanced features to obtain more accurate attention positions. Meanwhile, positional and linguistic information is constantly introduced in the PAM, so as to alleviate the linguistic insensitive drift problem. Compared with previous methods, we have a more concise structure and training strategy, while achieving better performance.

The main contributions of our work are as follows:
\begin{itemize}
    \item We are the first to point out the attention drift due to lack of linguistic information, which is called Linguistic Insensitive Drift (LID) problem, and propose a Cascade Position Attention mechanism to effectively handle the LID problem.
    \item We propose a Global Linguistic Reconstruction Module to reconstruct the features of each character by aggregating global linguistic information during the process of sequence modeling. The method does not introduce extra parameters.
    \item Our method achieves state-of-the-art performance while keeping a very low parameter quantity with a simple end-to-end training strategy.
\end{itemize}

\section{Related Work}
\subsection{Scene Text Recognition}
Scene Text Recognition (STR) has been a significant research term in computer vision. Early methods use a backbone and a sequence modeling network for feature extraction and use a Connectionist Temporal Classification (CTC)~\cite{CTC} decoder or attention decoder for prediction~\cite{seed,2dattn}. CTC-based decoder aims to maximize the probability of all the paths for final prediction, while attention-based decoder aims to localize the position of each character by attention mechanism. To further extract linguistic information of the visual predictions, SRN~\cite{SRN} proposed a language model to learn the relationship between each character. ABINet~\cite{abinet} further proposed a stronger bi-directional language model for autonomous linguistic modeling. We believe that a powerful recognizer must have the ability of contextual linguistic modeling, but explicit language models have a large number of parameters, which severely limits recognition efficiency.

Recently, the simplicity of model reasoning has been emphasized. Considering the CTC-based decoder has an advantage in speed while the attention-based decoder has an advantage in precision, GTC~\cite{gtc} used a powerful attention-based decoder to guide the training of a CTC-based decoder. MGP-STR~\cite{MGP} used ViT as the backbone to achieve high performance, which proved that the structure of ViT is applicable to STR. Further, SVTR~\cite{svtr} proposed a faster and more lightweight backbone for STR task. We think that in order to design a more powerful recognizer, the model must have linguistic modeling capability while keeping the structure simple. Thus, we propose the Global Linguistic Reconstruction Module, which can aggregate the contextual linguistic information during the process of sequence modeling. Such a design ensures the simplicity of the model.

\subsection{Attention Drift}
The visual attention drift problem in STR refers to the fact that when an attention-based decoder is used, the attention region of the decoder cannot be accurately aligned with the target character. \cite{focusingattention} first identified this problem and proposed a Focusing Attention Network (FAN) that is composed of an attention network for character recognition and a focusing network to adjust the attention drift. RobustScanner~\cite{robustscanner} deeply investigated the decoding process of the attention-based decoder and empirically find that a character-level sequence decoder utilizes not only context information but also positional information. They further suggested that the drowning of position information leads to attention drift problems. Using the above analysis, they solve this problem by a position enhancement branch to introduce position information. We further point out that attention drift comes from decoding different images with a linguistic insensitive query, which also lacks positional information. Based on this, we propose a Cascade Position Attention mechanism to solve this problem, which has a concise framework and does not introduce extra modules.

\section{Proposed Method}
In this section, we first detail the pipeline of proposed method in Sec. \ref{sec:pipeline}, and then we introduce Cascade Position Attention and Global Linguistic Reconstruction Module in Sec. \ref{sec:CPA} and Sec. \ref{sec:GLRM} respectively.

\subsection{Pipeline}\label{sec:pipeline}
The pipeline of our LPV is shown in Figure.\ref{fig:pipeline}. We can view the pipeline as two branches. Given an input image of size $H\times W\times 3$, the features $\mathbf{F}^i\in \mathbb{R}^{\frac{H}{4}\times \frac{W}{4}\times E}$ are obtained by the GLRM branch, which continuously enhances the features to obtain long-distance contextual linguistic information using proposed GLRM. Meanwhile, in the CPA branch, $\mathbf{F}^i$ are fed into the $i^{th}$ Position Attention Module (PAM) to get the attention map $\mathbf{A}^i$ and the feature $\mathbf{R}^i$ of each character. The CPA branch constantly rectifies the recognition results to alleviate linguistic insensitive drift using a linguistic-sensitive query. Note that the parameters in each PAM are \emph{NOT} shared.

\begin{figure}[!t]
\centering
\includegraphics[scale=0.5]{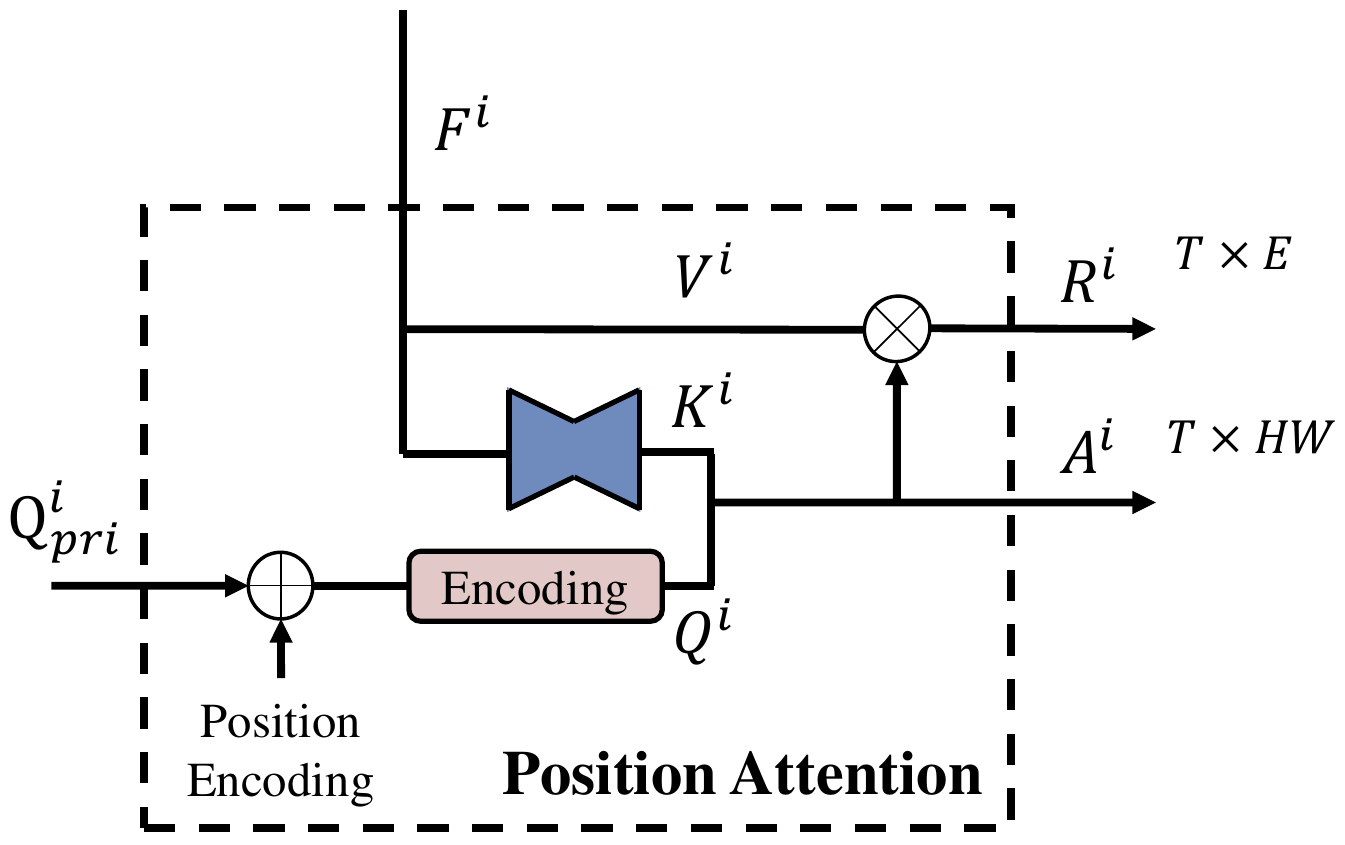}
\caption{The structure of Position Attention Module in the CPA. $\mathbf{Q}_{pri}$ is the prior query. 'Encoding' can be one FC layer.}
\label{fig:PAM}
\end{figure}

\subsection{Cascade Position Attention}\label{sec:CPA}
The Cascade Position Attention mechanism hierarchically optimizes the recognition result using the enhanced feature $\mathbf{F}^i$ in each stage and outputs the attention map of each character.

As shown in Figure. \ref{fig:PAM}, a cross-attention mechanism is utilized to transcribe visual features into character sequences. Specifically, the attention map $\mathbf{A}^i \in \mathbb{R}^{T\times \frac{HW}{16}}$ and the features $\mathbf{R}^i\in \mathbb{R}^{T\times E}$ of each character is calculated by the queries, keys, and values as Eq. \ref{eq:PAM}, where $T$ is the maximum length of the character sequence. The prediction results $\mathbf{Y}^i\in \mathbb{R}^{T\times C}$ can be further obtained by a classification head (e.g. FC Layer), where $C$ indicates the number of character classes. $\mathcal{P}(\cdot)$ is the classification head.

\begin{equation}\label{eq:PAM}
\begin{split}
    \mathbf{A}^i &= softmax(\mathbf{K}^i\mathbf{Q}^{i\top}/\sqrt{E}) \\
    \mathbf{R}^i &= \mathbf{A}^i\mathbf{V}^i \\
    \mathbf{Y}^i &= sofmax(\mathcal{P}(\mathbf{R}^i))
\end{split}
\end{equation}

Concretely, $\mathbf{K}^i=\mathcal{G}(\mathbf{F}^i)\in \mathbb{R}^{\frac{HW}{16}\times E}$, where $\mathcal{G}(\cdot)$ is implemented by a mini U-Net. $\mathbf{V}^i=\mathcal{H}(\mathbf{F}^i)\in \mathbb{R}^{\frac{HW}{16}\times E}$, where $\mathcal{H}(\cdot)$ is identity mapping. The most important, $\mathbf{Q}^i\in \mathbb{R}^{T\times E}$ is used to decode the position of each character and can be regarded as the priori of each character. Therefore, $\mathbf{Q}^i$ is generated by a given prior $\mathbf{Q}^i_{pri}$ and a position encoding $\mathbf{P}$ through the encoding layer (e.g. one FC layer). At the beginning of decoding, we do not know the specific information about the character so $\mathbf{Q}^0_{pri}$ is initialized to the $\mathbf{0}$ vectors. At the $i^{th}$ stage of decoding, $\mathbf{Q}^i_{pri}$ is set as the features of each character from the previous stage. The generation process of $\mathbf{Q}^i$ can be formalized as follows:

\begin{equation}
\begin{split}
    \mathbf{Q}^i_{pri} &= 
    \begin{cases}
        0, & i=0 \\
        \mathbf{R}^{i-1}, & otherwise \\
    \end{cases} \\
    \mathbf{Q}^i &= \mathcal{F}(\mathbf{Q}^i_{pri}+\mathbf{P})
\end{split}
\end{equation}
Where $\mathcal{F}(\cdot)$ is the encoding layer.

To deal with the problem of linguistic insensitive drift, on the one hand, through the continuous iteration of $\mathbf{Q}^i$, the network gradually gets a linguistic-sensitive query to decode the attention map. On the other hand, the positional information is constantly introduced by position encoding, which can enhance the positional sensitivity of the model.

\begin{figure}[!t]
\centering
\includegraphics[scale=0.35]{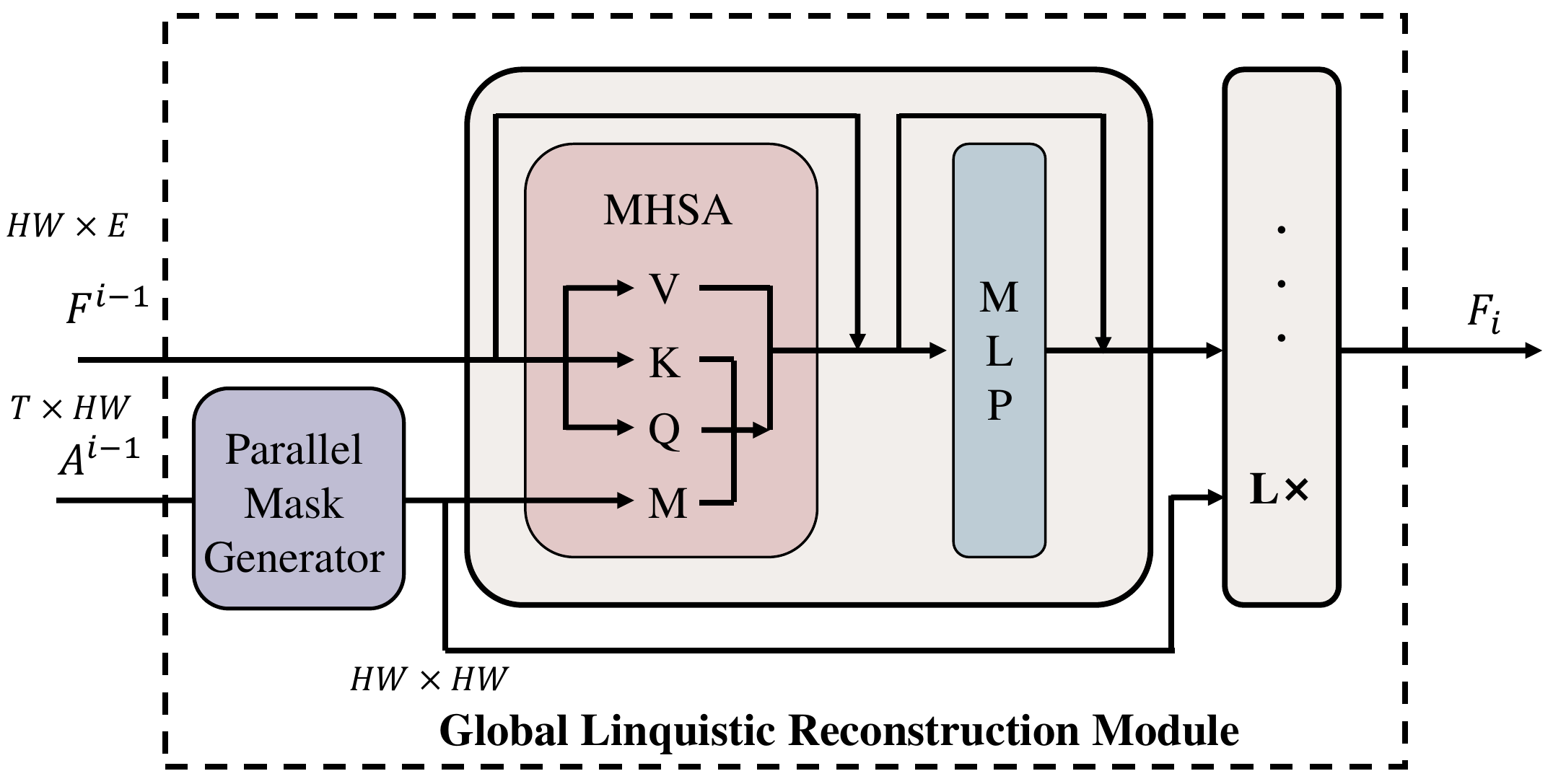}
\caption{The structure of Global Linguistic Reconstruction Module. GLRM consists of a Parallel Mask Generator and L $\times$ Masked Transformer Encoder.}
\label{fig:GLRM}
\end{figure}

\subsection{Global Linguistic Reconstruction Module}\label{sec:GLRM}
We argue that the input feature $\mathbf{F}^i$ of each stage can not be the same and it needs to be dynamically adjusted, e.g. sequence modeling. We will prove this inference in the ablation study. Therefore, it is necessary to add a sequence modeling network between stages but the simple sequence modeling network has no linguistic awareness, so we propose Global Linguistic Reconstruction Module to aggregate global linguistic information during sequence modeling without introducing extra parameters.

The details of GLRM is shown in Figure. \ref{fig:GLRM}, it takes the feature and attention map of the previous stage as inputs and outputs the enhanced feature of the current stage. GLRM contains two parts: Parallel Mask Generator (PMG) and Masked Transformer Encoder. The transformer encoder is proven to be effective for modeling long-range dependencies in recent computer vision tasks~\cite{transformerencoder1,transformerencoder2}, which can be used well for sequence modeling and contextual information aggregating. To guide the model learning linguistic knowledge, we design a novel way that reconstructs the features of each character by masking each character. To achieve this, PMG transforms the attention map $\mathbf{A}^{i-1}$ into a parallel mask $M\in \mathbb{R}^{\frac{HW}{16}\times \frac{HW}{16}}$ as Eq. \ref{eq:PMG}, where $\mathbf{U}(x)$ is the unit stage function which takes the value of 1 for $x\geq 0$ and 0 for $x<0$. $t$ is the threshold of foreground and background, which is set to 0.05 in our experiments. $\otimes$ is matrix multiplication.  

\begin{equation}\label{eq:PMG}
\begin{split}
    M^p &= \mathbf{U}(\mathbf{A}^{i-1}-t)^\top \otimes \mathbf{U}(\mathbf{A}^{i-1}-t)  \\
    M_{ij} &= 
    \begin{cases}
        -\infty, & M^p_{ij}>0 \\
        0, & otherwise \\
    \end{cases}
\end{split}
\end{equation}

Then, the attention operation inside multi-head self-attention blocks can be formalized as follows:

\begin{equation}\label{eq:MTE}
\begin{split}
    &[\mathbf{Q},\mathbf{K},\mathbf{V}]=\mathbf{F^{i-1}}\mathbf{W} \\
   &\mathbf{F}^i = softmax(\frac{\mathbf{Q}\mathbf{K}^\top}{\sqrt{E}}+M)\mathbf{V}
\end{split}
\end{equation}

By using such a mask, the tokens in one character can not see the tokens in the same character during the self-attention operation, which means that features within each character region are reconstructed from features other than that character. Benefiting from such a design, the visual features are gradually transformed into semantic-rich features during the cascade stage.

Compared with BERT~\cite{bert} and VisionLAN~\cite{visionlan}, though all approaches mask out the information in a certain time step, there are two differences: 1) BERT and VisionLAN mask the tokens of the input features, which leads to loss of origin features. But GLRM masks the tokens in the self-attention operation, which can ensure self-attention to model the global linguistic information. Meanwhile, the origin local features of each character are not lost due to the shortcut of the transformer encoder; 2) BERT and VisionLAN can only mask one character in a forward process, it can only guide the model to learn linguistic knowledge, but GLRM can mask all characters in a parallel way, which can reconstruct the features of each character and obtain an enhanced feature.

\begin{table*}[!t]
\centering
\begin{tabular}{@{}l|ccc|ccc|c|c|c@{}}
\toprule
\multirow{2}{*}{Method} & \multicolumn{3}{c|}{Regular}                  & \multicolumn{3}{c|}{Irregular}                & \multirow{2}{*}{AVG} & \#Params  &  Speed\\
                        & IC13          & SVT           & IIIT5k        & IC15          & SVTP          & CUTE          &                      & (M)    & (ms)  \\ \midrule
CRNN~\cite{CRNN}                    & 91.9          & 81.6          & 82.9          & 69.4          & 70.0          & 65.5          & 78.6                 & 8.3    & -  \\
ASTER~\cite{aster}                   & 91.8          & 89.5          & 93.4          & 76.1          & 78.5          & 79.5          & 86.7                 & 27.2   & -  \\
SEED~\cite{seed}                    & 92.8          & 89.6          & 93.8          & 80.0          & 81.4          & 83.6          & 88.3                 & -      & -  \\
RobustScanner~\cite{robustscanner}           & 94.8          & 88.1          & 95.3          & 77.1          & 79.5          & 90.3          & 88.4                 & -      & -  \\
TextScanner~\cite{textscanner}             & 92.9          & 90.1          & 93.9          & 79.4          & 84.3          & 83.3          & 88.5                 & -      & -  \\
SRN~\cite{SRN}                     & 95.5          & 91.5          & 94.8          & 82.7          & 85.1          & 87.8          & 90.4                 & 54.7   & -  \\
VisionLAN~\cite{visionlan}               & 95.7          & 91.7          & 95.8          & 83.7          & 86.0          & 88.5          & 91.2                 & 32.8   & 21.73  \\
ABINet~\cite{abinet}                  & 97.4          & 93.5          & 96.2          & 86.0          & 89.3          & 89.2          & 92.3                 & 36.7   & 46.86  \\ \midrule
MGP-Small~\cite{MGP}               & 96.4          & 93.5          & 95.3          & 86.1          & 87.3          & 87.9          & 92.0                 & 52.6   & -  \\
MGP-Base~\cite{MGP}                & \underline{97.3}          & \textbf{94.7}          & 96.4          & \underline{87.2}          & \textbf{91.0}          & 90.3          & 93.3                 & 148.0  & -  \\ \midrule
SVTR-Tiny~\cite{svtr}               & 96.3          & 91.6          & 94.4          & 84.1          & 85.4          & 88.2          & 90.8                 & 6.03   & 4.11  \\
SVTR-Small~\cite{svtr}              & 95.7          & 93.0          & 95.0          & 84.7          & 87.9          & 92.0          & 91.6                 & 10.3   & 4.81  \\
SVTR-Base~\cite{svtr}               & 97.1          & 91.5          & 96.0          & 85.2          & 89.9          & 91.7          & 92.3                 & 24.6   & 5.80  \\ \midrule
LPV-Tiny (Ours)               & 96.7          & 92.9          & 96.3          & 86.4          & 86.7          & 90.6          & 92.5                 & 8.11   & 5.17  \\
LPV-Small (Ours)             & 96.8          & 93.7          & \underline{96.7}          & 87.1          & 89.8          & \underline{92.4}          & \underline{93.3}                 & 13.99  & 5.77  \\
LPV-Base (Ours)              & \textbf{97.6} & \underline{94.6} & \textbf{97.3} & \textbf{87.5} & \underline{90.9} & \textbf{94.8} & \textbf{94.0}        & 35.13  & 7.41  \\ \bottomrule
\end{tabular}
\caption{Results on IC13, SVT, IIIT5K, IC15, SVTP and CUTE datasets. Following~\protect\cite{abinet,MGP}, all the results are under NONE lexicon. The speed is the inference time on one NVIDIA 2080Ti GPU averaged over 1000 English image text.}
\label{tab:performance}
\end{table*}

\subsection{Training Objective}
The final objective function of the proposed method is formulated in Eq. \ref{eq:loss}. $N$ is the number of cascade stages and $\mathbf{Y}^i$ is the prediction at the $i^{th}$ stage. $g_t$ is the ground truth. $T$ is the max length of the character sequence which we set to 25 in our experiments.

\begin{equation}\label{eq:loss}
   L = -\frac{1}{NT}\sum_{i=0}^{N}\sum_{j=0}^{T} log(P(\mathbf{Y}^i|g_t))
\end{equation}

\section{Experiment}
\subsection{Datasets}
For fair comparison, we conduct experiments following the setup of \cite{MGP,abinet}. We use MJSynth~\cite{MJ1,MJ2} and SynthText~\cite{ST} as training data and they contain 9M and 7M synthetic text images respectively. The performance is evaluated on 6 benchmarks containing IIIT 5K-Words (IIIT5K)~\cite{IIIT}, ICDAR2013 (IC13)~\cite{IC13}, ICDAR2015 (IC15)~\cite{IC15}, Street View Text (SVT)~\cite{SVT}, Street View Text-Perspective (SVTP)~\cite{SVTP} and CUTE80 (CUTE)~\cite{CUTE}. Details of the above 6 datasets can be found in previous works \cite{MGP,abinet}.

\subsection{Implementation Details}
We use the backbone proposed in SVTR \cite{svtr} as our backbone due to its impressive performance in STR. Particularly, to use the attention-based decoder, we change the stride of the merging module to 1 and remove the final mixing head to obtain visual features at 1/4 resolution. The image size is set to 100 × 32. Following the most recent works~\cite{MGP,benchmark}, for fair comparison, we use the same code framework and data augmentation. We conduct the experiments on 4 NVIDIA 3090 GPUs with batch size 384. The vocabulary size C of character classification head is set to 38, including 0 - 9, a - z, [PAD] for padding symbol and [EOS] for ending symbol.

The network is trained end-to-end using Adam~\cite{adam} optimizer of initial learning rate 1e-4 and the learning rate is decayed to 1e-5 after six epochs. We trained a total of 20 epochs. The first 10 epochs do not use the mask we proposed in GLRM so that the position attention can obtain a relatively accurate attention map. The last 10 epochs add the mask for finetune so that the network can learn contextual linguistic knowledge.

\subsection{Comparisons with State-of-the-Arts}
We compare our method with previous state-of-the-art methods on 6 benchmarks in Table. \ref{tab:performance}. Our model shows significant performance in both regular (IC13, SVT and IIIT5K) and irregular (IC15, SVTP and CUTE) datasets while keeping a very low parameter quantity. Notably, LPV-Tiny has already outperformed most of the state-of-the-art methods with only 8.11M parameters while LPV-Small and LPV-Base obtain the performance of 93.3\% and 94.0\% with only 13.99M and 35.13M parameters respectively. For inference time, LPV-Tiny only needs 5.17ms, which is faster than most of the existing methods.

Many previous works, such as SRN~\cite{SRN}, ABINet~\cite{abinet}, VisionLAN~\cite{visionlan}, and MGP~\cite{MGP} tried to introduce linguistic knowledge to assist recognition. Compared to them, LPV shows the best performance on all datasets. This result implies that our cascade position attention mechanism and GLRM are effective. Additionally, compared with SVTR, our tiny, small, and base model obtains 1.6\%, 1.7\%, and 1.7\% improvement respectively.

\begin{table}[!t]
\centering
\begin{tabular}{@{}cccccc@{}}
\toprule
\multirow{2}{*}{N}        & IC13            & SVT             & IIIT5k          & \multirow{2}{*}{AVG}             & Params                \\
                          & IC15            & SVTP            & CUTE            &                                  & (M)                   \\ \midrule
\multirow{2}{*}{-} & 96.3            & 91.6            & 94.4            & \multirow{2}{*}{90.9}            & \multirow{2}{*}{6.03} \\
                          & 84.1            & 85.4            & 88.2            &                                  &                       \\ \midrule
\multirow{2}{*}{1}        & 96.616          & 91.808          & 95.733          & \multirow{2}{*}{91.336}          & \multirow{2}{*}{4.39} \\
                          & 84.705          & 86.512          & 90.625          &                                  &                       \\ \midrule
\multirow{2}{*}{2}        & 96.033          & 92.581          & 95.833          & \multirow{2}{*}{92.150}          & \multirow{2}{*}{6.25} \\
                          & 85.864          & \textbf{87.752} & 90.625          &                                  &                       \\ \midrule
\multirow{2}{*}{3}        & \textbf{96.733} & \textbf{92.890} & \textbf{96.300} & \multirow{2}{*}{\textbf{92.481}} & \multirow{2}{*}{8.11} \\
                          & \textbf{86.361} & 86.667          & \textbf{90.625} &                                  &                       \\ \bottomrule
\end{tabular}
\caption{Ablation study of Cascade Position Attention (CPA) mechnaism in LPV-Tiny, N is the number of stages in CPA. The first row is the baseline with a CTC-based decoder.}
\label{tab:CPA}
\end{table}

\subsection{Ablation Study}
\subsubsection{The Effectiveness of Cascade Position Attention}
We propose the Cascade Position Attention (CPA) mechanism to alleviate the linguistic insensitive drift problem. To prove the effectiveness of CPA, we perform ablation from two aspects.

From the aspect of model performance, we conduct several experiments to evaluate the effect of the number of stages $N$ in Table. \ref{tab:CPA}. Especially, $N=1$ means no extra stage to optimize the recognition result. The first row of Table \ref{tab:CPA} is the baseline with a CTC-based decoder instead of attention-based decoder. From the statistics we can conclude: 1) Our model with 3 stages outperforms 1.581\% more improvement. 2) Due to the concise structure of the hierarchical optimization strategy, the increase of stages will result in great gains in average accuracy while little increase in parameter quantity. 3) As the number of stages increases gradually, performance improvement is limited. For the trade-off between parameter quantity and accuracy, we choose 3 stages in our model.

\begin{figure*}[!t]
\centering
\includegraphics[scale=0.5]{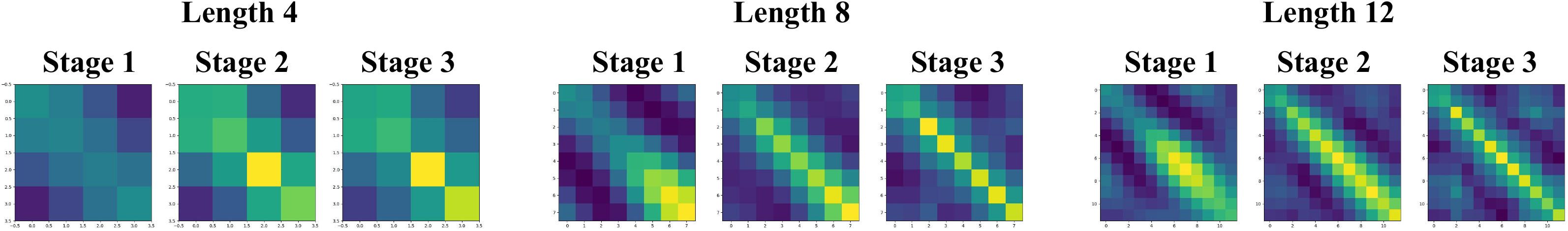}
\caption{The visualization of the similarity between the queries at different positions in LPV-Tiny.}
\label{fig:sim}
\end{figure*}

From the aspect of attention drift, we further visualize it. In the position attention mechanism, the query guides the decoder to find the position of each character. As described in Sec. \ref{sec:introduction}, the high similarity of queries between neighboring locations leads to the problem of attention drift. Based on LPV-Tiny, we visualize the similarity between query vectors at different positions in Figure. \ref{fig:sim}. Due to stages 2 and 3 having a linguistic-sensitive query that is different when inputting different images, we calculate the average similarity with all images in IC15~\cite{IC15} of each sequence length. As shown in Figure \ref{fig:sim}, the query in the first stage has no position and linguistic prior about the input image so it does not have a centralized similarity. In stages 2 and 3, the queries consist of a linguistic prior query $\mathbf{Q}_{pri}$, and the position encoding is introduced again to enhance position sensitivity. Therefore, the similarity is centered in the diagonal, which means the position of each character is more certain. When decoding characters, the feature similarity of neighboring characters is reduced, so the attention drift is mitigated. Note that the similarity in stage 3 is more concentrated than that in stage 2 due to the stronger prior and more position information. Additionally, the difference is even more pronounced with long text, because attention drift is more likely to occur in the case of long text.

\begin{table}[!t]
\centering
\begin{tabular}{@{}c|cccccc@{}}
\toprule
\multirow{2}{*}{Str}   & \multirow{2}{*}{Feat} & \multirow{2}{*}{Mask} & IC13            & SVT             & IIIT5k          & \multirow{2}{*}{AVG}             \\
                       &                       &                       & IC15            & SVTP            & CUTE            &                                  \\ \midrule
\multirow{6}{*}{Tiny}  & \multirow{2}{*}{D}    & \multirow{2}{*}{\XSolidBrush}    & 95.683          & 92.736          & 95.800          & \multirow{2}{*}{92.012}          \\
                       &                       &                       & 85.588          & \textbf{87.597} & 90.278          &                                  \\ \cmidrule(l){2-7} 
                       & \multirow{2}{*}{D}    & \multirow{2}{*}{\Checkmark}    & \textbf{96.733} & \textbf{92.890} & \textbf{96.300} & \multirow{2}{*}{\textbf{92.481}} \\
                       &                       &                       & \textbf{86.361} & 86.667          & 90.625          &                                  \\ \cmidrule(l){2-7} 
                       & \multirow{2}{*}{F}    & \multirow{2}{*}{\XSolidBrush}    & 95.683          & 91.499          & 95.433          & \multirow{2}{*}{91.763}          \\
                       &                       &                       & 85.919          & 86.512          & \textbf{90.972} &                                  \\ \midrule
\multirow{4}{*}{Small} & \multirow{2}{*}{D}    & \multirow{2}{*}{\XSolidBrush}    & 96.383 & 93.045          & 96.533          & \multirow{2}{*}{92.701}          \\
                       &                       &                       & 86.582          & 88.217          & 89.583          &                                  \\ \cmidrule(l){2-7} 
                       & \multirow{2}{*}{D}    & \multirow{2}{*}{\Checkmark}    & \textbf{96.849} & \textbf{93.663} & \textbf{96.663} & \multirow{2}{*}{\textbf{93.240}} \\
                       &                       &                       & \textbf{87.134} & \textbf{89.767} & \textbf{92.361} &                                  \\ \bottomrule
\end{tabular}
\caption{Ablation study of GLRM, in the column of Feat, 'D' means dynamic feature in the sequence modeling and 'F' means fixed feature. Mask indicates if use the mask we peoposed.}
\label{tab:GLRM}
\end{table}

\begin{table}[!t]
\centering
\begin{tabular}{@{}cccccc@{}}
\toprule
\multirow{2}{*}{L} & IC13            & SVT             & IIIT5k          & \multirow{2}{*}{AVG}             & Params                \\
                   & IC15            & SVTP            & CUTE            &                                  & (M)                   \\ \midrule
\multirow{2}{*}{1} & 96.616          & 91.808          & 95.733          & \multirow{2}{*}{91.708}          & \multirow{2}{*}{6.53} \\
                   & 84.705          & 86.512          & \textbf{90.625} &                                  &                       \\ \midrule
\multirow{2}{*}{2} & 96.733          & 92.890          & 96.300          & \multirow{2}{*}{92.481}          & \multirow{2}{*}{8.11} \\
                   & \textbf{86.361} & 86.667          & 90.625          &                                  &                       \\ \midrule
\multirow{2}{*}{3} & \textbf{97.083} & \textbf{92.890} & \textbf{96.500} & \multirow{2}{*}{\textbf{92.632}} & \multirow{2}{*}{9.69} \\
                   & 86.140          & \textbf{87.907} & 89.931          &                                  &                       \\ \bottomrule
\end{tabular}
\caption{Ablation study of the layer number of GLRM. L is the layer number of GLRM. Experiments were performed on LPV-Tiny.}
\label{tab:layer}
\end{table}

\begin{figure*}[!t]
\centering
\includegraphics[scale=0.5]{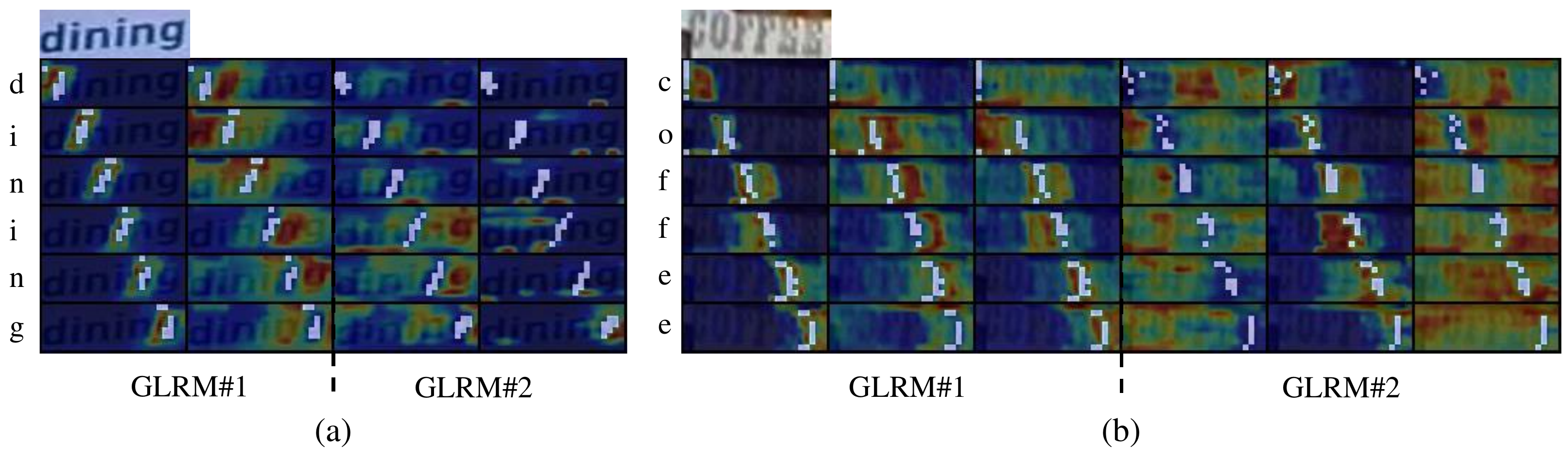}
\caption{The average attention map of the pixels in the area masked and the area masked is shown in white. (a) LPV-Tiny with 2 layers in each GLRM. (b) LPV-Small with 3 layers in each GLRM.}
\label{fig:GLRMattn}
\end{figure*}

\subsubsection{The Effectiveness of GLRM}
As described in Sec. \ref{sec:GLRM}, we argue that the input features $\mathbf{F}^i$ of each stage can not be the same and needs to be dynamically adjusted, so a sequence modeling network is necessary. To prove this inference, we first use a simple transformer encoder as the sequence modeling network to obtain dynamic features. For fair comparison, we place the transformer encoder before the CPA decoder to fix the features and keep the same parameter quantity. As shown in Table. \ref{tab:GLRM}, the performance is not good (91.763\% vs 92.012\%) when we place the transformer layer before the CPA and input the same features into each stage.

Furthermore, to acquire linguistic knowledge, we propose GLRM as the sequence modeling network which uses a parallel mask to enhance the feature and obtain the contextual linguistic information. As shown in Table. \ref{tab:GLRM}, for LPV-Tiny and LPV-Small, the proposed mask obtains 0.469\%  and 0.539\% improvement on average accuracy respectively.

\subsubsection{The Layer Number of GLRM}
Our GLRM consists of a Parallel Mask Generator and $L\times$ Masked Transformer Encoder. To determine the number of layers L, we conduct several experiments. From the results in Table. \ref{tab:layer} we can observe: 1) More layers in GLRM can provide stronger contextual modeling capacity and obtain higher performance. 2) We can obtain 0.69\% improvement when L increases from 1 to 2, which is greatly larger than 0.234\% when L goes from 2 to 3. That is because the masked transformer encoder can only model the area around each character at a shallow layer, and gradually model the global feature as it moves deeper. This conjecture can be verified by the visualization of the attention map in the masked transformer encoder. We calculate the average attention map of the pixels in the area masked and show the visualization in Figure. \ref{fig:GLRMattn}. From the attention map, we can find that the attention in the first layer of the first GLRM is limited around each character because there is no global feature in the input. When it goes deep, the attention area goes global. 

Finally, considering the total parameter quantity, we set $L$ to 2 in LPV-Tiny and 3 in LPV-Small and LPV-Base.

\subsection{The Qualitative Analysis}
\subsubsection{GLRM in Subword Perception}
From the visualization in Figure. \ref{fig:GLRMattn}, we can further analyze the attention area. As we all know, there are some sub-words that occur frequently in words (e.g. 'ing', 'pri', 'mer', 'tion', etc). Such knowledge can assist the model to obtain a more accurate result when the visual clue is confused. Our GLRM guides the model to reconstruct the feature of each character using the feature of other characters so it will be sensitive to the sub-words. As shown in Figure \ref{fig:GLRMattn} (a), the sub-words 'din' and 'ing' pay attention to themselves individually. This demonstrates the ability of our GLRM to learn contextual linguistic knowledge.

\subsubsection{Lguistic Insensitive Drift Problem}
\begin{figure}[!t]
\centering
\includegraphics[scale=0.28]{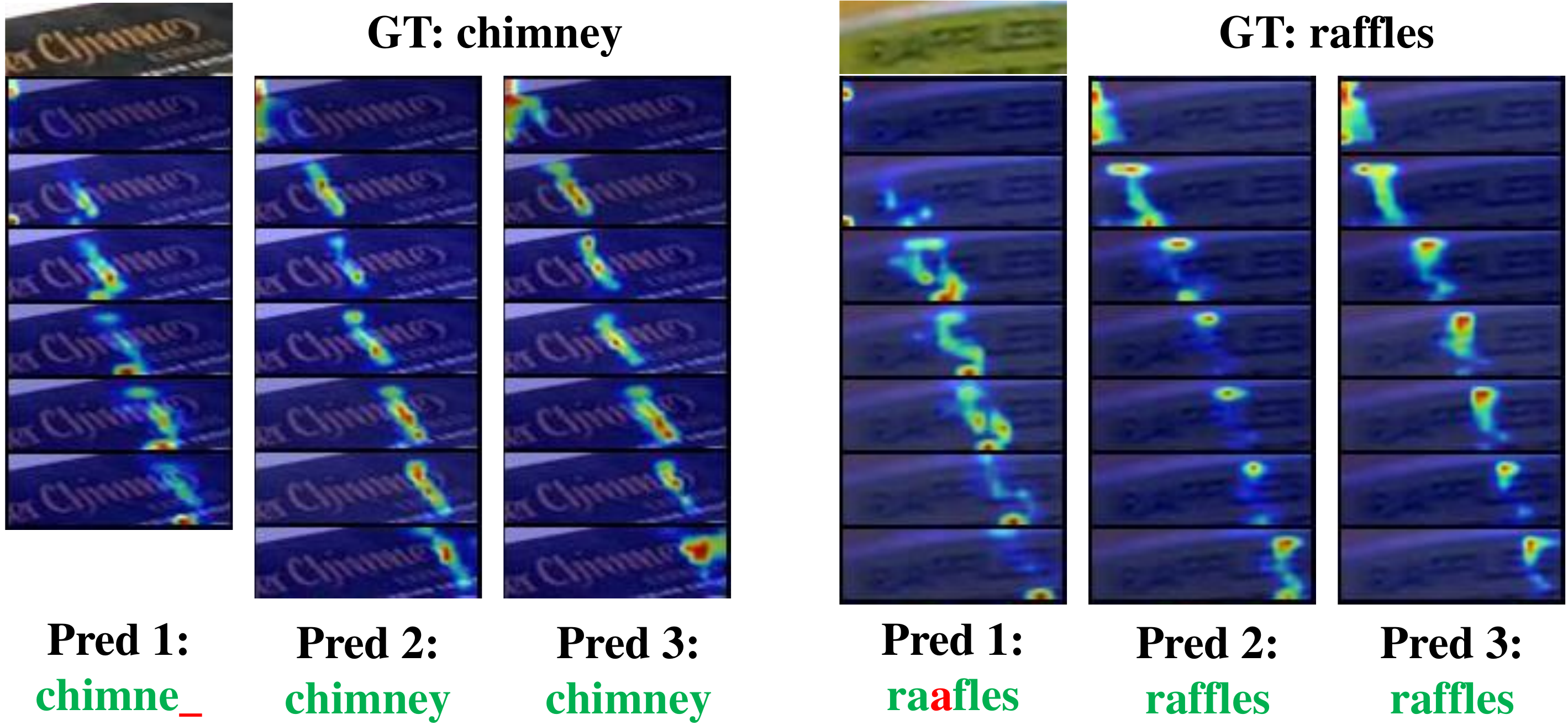}
\caption{The visualization of attention map in each stage of CPA. Experiments were performed on LPV-Small. The model has three stages in CPA so there are three predict results for each input.}
\label{fig:attn}
\end{figure}

Figure. \ref{fig:attn} shows some sample cases of attention drift being corrected. For each input image, LPV can get three stages of recognition results: one preliminary result and two correction results. From the attention map, we can observe that if the $1^{st}$ stage gets a drift result, the remaining stages have the ability to correct benefiting from the linguistic-sensitive query in CPA.

\section{Conclusion}
This paper first notices the Linguistic Insensitive Drift (LID) problem and analyzes the linguistic perception of the model. To find an efficient and accurate method, LPV is proposed to enhance the linguistic information of both query and feature (Linguistic More). To be specific, LPV introduces CPA to obtain an accurate attention map by using linguistic-sensitive query instead of visual query, and designs GLRM to aggregate the global linguistic information to enhance the visual feature. Compared with previous methods, our LPV is able to take a further step toward efficient and accurate recognition, which obtains dominant recognition performance while maintaining a concise pipeline. We believe that LPV will inspire recent works in simple network design and efficient linguistic perception, and we will further explore its potential in the future.

\section*{Acknowledgments}
This work is supported by the National Key Research and Development Program of China (2022YFB3104700), the National Nature Science Foundation of China (62121002, 62022076, U1936210, 62232006).

\bibliographystyle{named}
\bibliography{ijcai23}

\end{document}